\documentclass{article}

\usepackage[english]{babel}

\usepackage[letterpaper,top=2cm,bottom=2cm,left=3cm,right=3cm,marginparwidth=1.75cm]{geometry}

\usepackage{amsmath}
\usepackage{graphicx}
\usepackage[colorlinks=true, allcolors=blue]{hyperref}

\title{Heuristic Vision Pre-Training with Self-Supervised and Supervised Multi-Task Learning}
\author{Zhiming Qian}

\begin{document}
\maketitle

\begin{abstract}
To mimic human vision with the way of recognizing the diverse and open world, foundation vision models are much critical. While recent techniques of self-supervised learning show the promising potentiality of this mission, we argue that signals from labelled data are also important for common-sense recognition, and properly chosen pre-text tasks can facilitate the efficiency of vision representation learning. To this end, we propose a novel pre-training framework by adopting both self-supervised and supervised visual pre-text tasks in a multi-task manner. Specifically, given an image, we take a heuristic way by considering its intrinsic style properties, inside objects with their locations and correlations, and how it looks like in 3D space for basic visual understanding. However, large-scale object bounding boxes and correlations are usually hard to achieve. Alternatively, we develop a hybrid method by leveraging both multi-label classification and self-supervised learning.  On the one hand,  under the multi-label supervision, the pre-trained model can explore the detailed information of an image, e.g., image types, objects, and part of semantic relations. On the other hand,  self-supervised learning tasks, with respect to Masked Image Modeling (MIM)  and  contrastive learning, can help the model learn pixel details and patch correlations. Results show that our pre-trained models can deliver results on par with or better than state-of-the-art (SOTA) results on multiple visual tasks. For example, with a vanilla Swin-B backbone, we achieve 85.3\% top-1 accuracy on ImageNet-1K classification, 47.9 box AP on COCO object detection for Mask R-CNN, and 50.6 mIoU on ADE-20K semantic segmentation when using Upernet. The performance shows the ability of our vision foundation model to serve general purpose vision tasks. 

\end{abstract}

\section{Introduction}

To learn the intrinsic universal knowledge of visual world,  pre-training models are motivated to learn fundamental representations to support a broad range of downstream tasks, similar to what humans would do ~\cite{yuan2021florence}. One milestone for the pre-training issue is the introduction of transfer learning \cite{pan2009survey}, which formalizes a two-stage learning framework: a pre-training stage to capture knowledge from one or more source tasks, and a fine-tuning stage to transfer the captured knowledge to target tasks. Owing to the wealth of knowledge obtained in the pre-training stage, the fine-tuning stage can enable models to well handle target tasks with limited samples. Specifically, supervised pre-training with image classification on ImageNet \cite{deng2009imagenet} has driven the progress in solving many computer vision tasks in the past few years, such as image classification \cite{dosovitskiy2020image}\cite{liu2021swin}, object detection \cite{he2017mask}\cite{cai2018cascade} and semantic segmentation \cite{kirillov2019panoptic}\cite{xiao2018unified}. Recently, study in self-supervised pre-training shows that it can generalize well for specific downstream tasks by taking ingenious strategies of many self-supervised objectives, such as contrastive learning \cite{chen2020simple}\cite{chen2021empirical}\cite{caron2021emerging}\cite{caron2020swav} and Masked Image Modeling (MIM) \cite{bao2021beit}\cite{he2021masked}\cite{xie2021simmim}\cite{dong2021peco}. 

\begin{figure}
\centering
\includegraphics[width=0.5\textwidth]{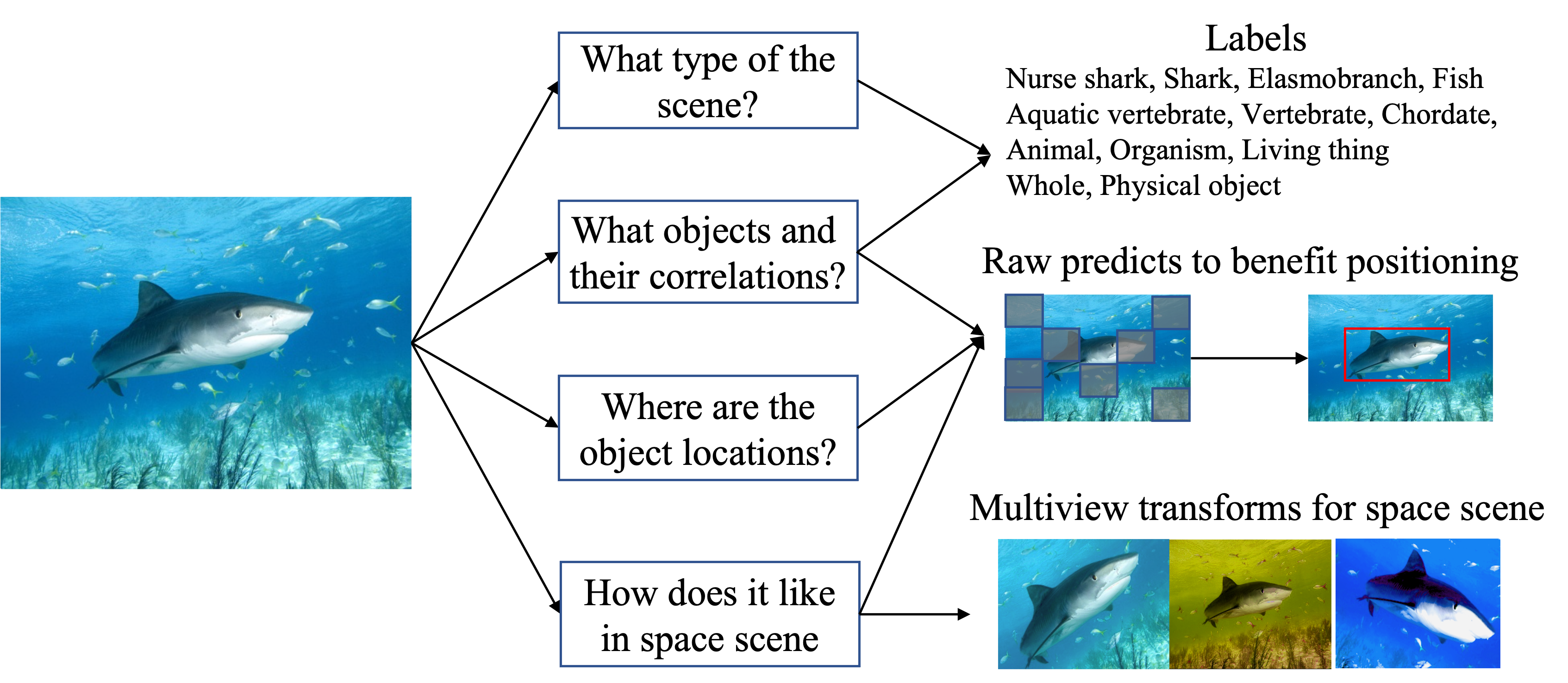}
\caption{\label{fig:illustration}An illustration of a few heuristic insights. For an image, we understand its visual content by simultaneously perceiving scene properties, inside objects and their correlations, motivating us to learn visual representations with the relevant tasks.}
\end{figure}

To investigate representation between supervised and self-supervised methods, Grigg et al. \cite{grigg2021self} recently find that supervised and self-supervised methods learn similar intermediate representations through dissimilar means, but diverge rapidly in the final few layers. The similarity indicates a shared set of primitives, and the divergence is probably caused by the layers strongly to the distinct learning objectives. Furthermore, taking both weak supervision of image labels and self-supervision of each single modality, multi-model methods   \cite{yuan2021florence}\cite{li2021albef} can achieve much competitive results on the authoritative visual challenge tasks. Besides, the absolute model size for current vision models is just able to reach about 1-2 billion parameters \cite{liu2021swinv2}, resulting in the fact that the need of large-scale unlabelled data for self-supervised learning is not urgent \cite{el2021large}. Based on these views, we employ a large-scale multi-label dataset for both supervised and self-supervised learning, and design a Heuristic Vision Pre-training method with Multi-Task Learning (HVP-MTL) by combining both supervised multi-label classification and self-supervised objectives. We believe that the open large-scale supervised datasets can currently make good generalization performance, which has been proved with the advanced work \cite{dosovitskiy2020image} based on JFT-300M \cite{sun2017revisiting}. Specifically, the Tencent-ML dataset \cite{tencent-ml-images-2019} is employed. Then, with the purpose of learning fundamental representations, we first set a few heuristic problems. As seen in Figure \ref{fig:illustration}, given an image, it is natural to ask some questions for understanding, such as which type of the image is it, what are the objects and their correlations, where are these objects, and how does it like in 3D space. To cope with these problems, we propose a novel framework by taking supervised and self-supervised tasks, including multi-label classification, reconstruction with masked images, and embedding alignment with different image views. The relations between the above heuristic problems and pre-text tasks are illustrated in Figure \ref{fig:illustration}. Our contributions are summarized as follows:

\begin{itemize}
\item We propose a unified framework for multi-task learning by setting a few heuristic pre-text tasks, with the purpose of learning basic visual representations.  Supervised pre-text tasks can usually achieve sustainable gain with the increasing of the dataset size, and self-supervised pre-text tasks are class-agnostic and promising for learning fundamental structures. Combined both supervised and self-supervised pre-text tasks in a heuristic way, we can learn more consistent representations with human beings.
\item For supervised learning, we adopt multi-label classification, and employ momentum distillation\cite{li2021albef} for label denoising. To solve the label imbalance problem, we develop a novel weighted asymmetric loss \cite{ridnik2021asymmetric} for multi-label classification. 
\item For self-supervised learning, we use Masked Image Modeling (MIM)\cite{xie2021simmim} for implicitly infer intrinsic objects with their locations and correlations, and employ contrastive learning for embedding alignment with different image views, which can benefit scene understanding in 3D space. To make contrastive learning more efficient and stable, we take online clustering with SWaV \cite{caron2020swav}, and impose a layer truncation to solve the collapse problem of the exponential computation when using Sinkhorn-Knopp \cite{cuturi2013sinkhorn}.
\end{itemize}

\section{Related Work}
\subsection{Multi-task learning} 
Multi-task learning can bring more insightful interpretation for learning features, but might suffer from negative transfer due to task conflicts \cite{nassif2020multitask}. To overcome this, works such as GRAD-CAM \cite{selvaraju2017grad} proposes techniques that provide visual explanations for decisions made by a model to make them more transparent and explainable. Then, multi-model methods, such as ALBEF \cite{li2021albef} and Florence \cite{yuan2021florence}, utilize both weak supervision of image descriptions and self-supervision of each single modality for pre-training, achieving a great success on downstream visual tasks. In our study, we use multi-task learning by setting a few heuristic pre-text tasks, with the purpose of  learning shared features among these prompting pre-text tasks and finding the intrinsic image representation.

\subsection{Multi-label classification} 
Multi-label classification are more natural descriptions for images, and can tell the image types, properties, inside objects, or even the correlations among objects. For its nature of multiple labels on one image, the co-occurrence of concepts in a large-scale dataset could be mined as prior knowledge for subsequent classification. A key characteristic of multi-label classification is the inherent positive-negative imbalance created when the overall number of labels is large \cite{tencent-ml-images-2019}. To address this issue, a few work suggests using a dedicated loss function to statically handle the imbalance, such as distribution-balanced loss \cite{wu2020distribution}, focal loss \cite{lin2017focal}, asymmetric loss \cite{ridnik2021asymmetric}. Another key characteristic is label correlation, graph-based methods \cite{chen2019learning} and class-aware maps\cite{chen2019multi}\cite{ye2020attention} are employed to represent the relationship of labels. While modeling label correlations can introduce additional gains in multi-label classification, it is also arguable that it may learn spurious correlations when the label statistics are insufficient. Rather than using graph,  the work in \cite{liu2021query2label} leads the network to focus on regions of interest for implicitly capturing label relationships by introducing a Transformer decoder. However, few work focuses on the intrinsic noising problem in the multi-class dataset. In our work, the Transformer decoder is applied with a novel weighted asymmetric loss, and we employ momentum distillation \cite{li2021albef} for label denoising.

\subsection{Self-supervised learning}
Self-supervised learning has attracted increasing attention over the past few years, as deep learning networks become more and more data-hungry and it is impossible to label everything in the world. There are two main categories to alleviate this issue, w.r.t. contrastive and generative. Contrastive learning is a discriminative approach that aims at grouping similar samples to be closer and dissimilar samples to be far from each other. By using a noise contrastive estimator (NCE) \cite{oord2018representation} to compare instances instead of classifying them, dealing with a large number of images simultaneously is usually required for good performance. In practice, this requires large batches \cite{chen2020simple} or memory banks \cite{chen2021empirical}. In short, contrastive-based methods heavily depend on the strong data augmentation and effective negatives sampling. To alleviate this, several variants allow automatic grouping of instances in the form of clustering \cite{caron2020swav}. Here, we take a robust online clustering method for learning similarity from different views, with the purpose of pursuing memory efficiency and visual coherence.

The other recent resurgent field is generative self-supervised learning \cite{bao2021beit}\cite{he2021masked}, training an encoder and a decoder under the objective of reconstruction loss, aiming at recovering the corrupted or masked input, which has yielded the most successful frameworks in NLP \cite{devlin2018bert}.  Recently, BEiT \cite{bao2021beit}  proposes a pre-text task of MIM by recovering the original visual tokens based on the corrupted image patches. Then, MAE \cite{he2021masked} reconstructs pixels with an asymmetric encoder-decoder architecture by masking a high proportion of the input image. More recently, PeCo \cite{dong2021peco} refine the visual codebooks, and SimMIM \cite{xie2021simmim} further study the influence of patch masking strategies. In this work, we take the pre-text task of MIM based on Transformers by directly learning from raw pixels to avoid the information loss.

\section{Method}
In this section, we first introduce the overview of our framework in Section \ref{sec_3_1}. Then, the pre-training objectives are delineated in Section \ref{sec_3_2}. In the end, we describe the pre-training dataset and implementation details in Section \ref{sec_3_3}.

\begin{figure*}
\centering
\includegraphics[width=0.9\textwidth, scale=1.0]{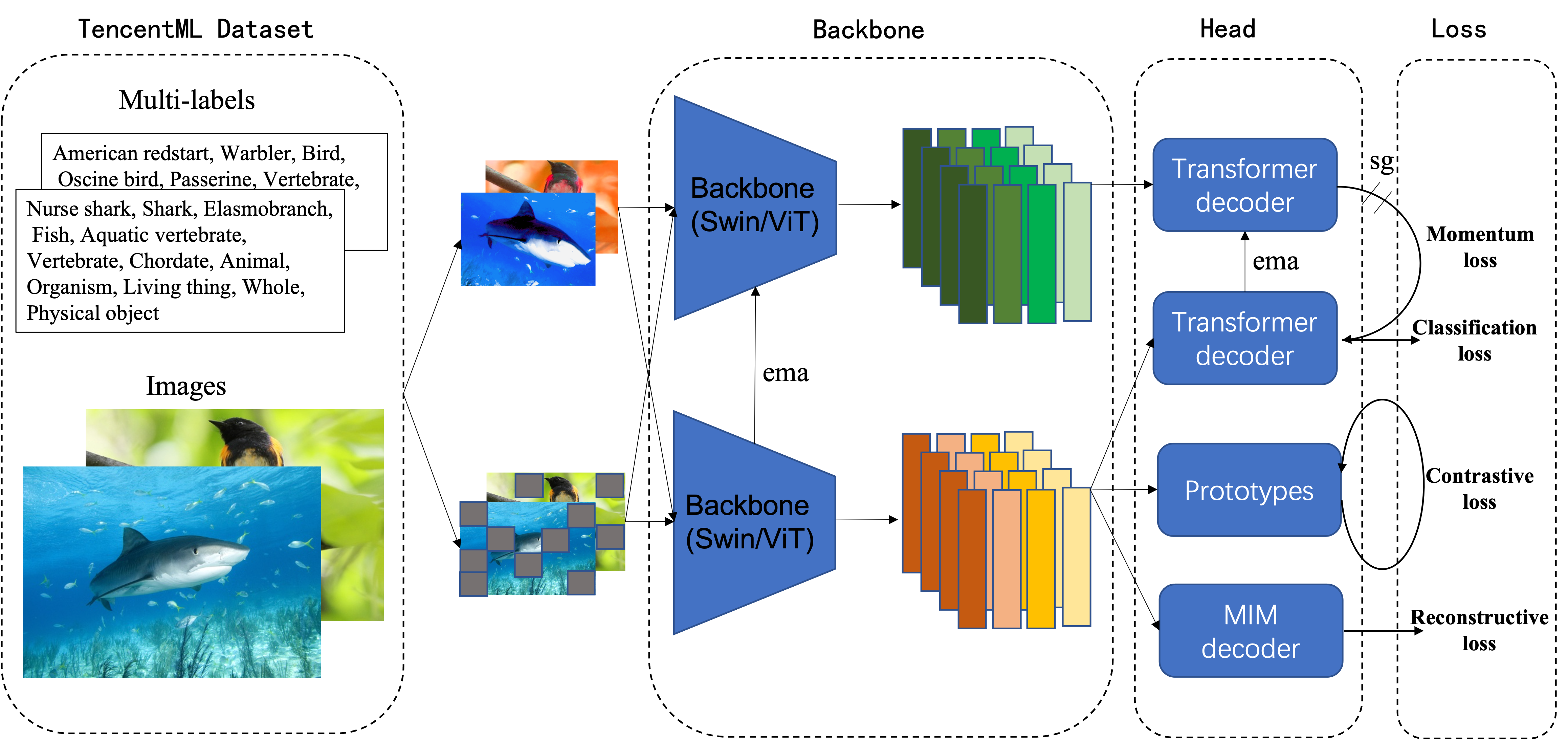}
\caption{\label{fig:overview}The pipeline of HVP-MTL. A Transformer backbone is first employed to encode image to a feature map. Then, we take several decoders for tasks of multi-label classification, MIM, contrastive learning and momentum distillation. Here, $sg$ represents stop gradient, and EMA is used for update parameters of the teacher network.}
\end{figure*}

\subsection{\label{sec_3_1}Overall architecture}
As illustrated in Figure \ref{fig:overview}, an image is first transformed into different views by conducting a few augmentations, such as color jittering, random cropping, patch masking, random rotation and so on. Then, an image encoder with the Swin \cite{liu2021swin} or ViT \cite{dosovitskiy2020image} backbone is employed to generate the feature map, which is usually the output of the last stage of the backbone. Furthermore, several head decoders with different losses are introduced for heuristic representation learning, including a Transformer decoder for multi-label classification, a clustering decoder with prototypes \cite{caron2020swav} for contrastive learning, and a MIM decoder for reconstructive learning. Besides, momentum distillation is employed for label denoising.

\subsection{\label{sec_3_2}Pre-training objectives}

\textbf{Transformer decoder for multi-label classification.} Given an image $x \in \mathbf{R}^{H_0 \times W_0 \times 3}$ as input, we extract its spatial features $\mathcal{F}_0 \in \mathbf{R}^{H \times W \times d_0}$ using the backbone, where $H_0 \times W_0, H \times W$ represent the height and weight of the input image and the feature map respectively, and $d_0$ denotes the dimension of features. After that, we add a linear projection layer to project the features from dimension $d_0$ to $d$ to match with the desired query dimension in the following Transformer decoder, and reshape the projected features to be $\mathcal{F} \in \mathbf{R}^{H \times W \times d}$. Finally, we use label embeddings as queries $Q_0 \in \mathbf{R}^{C \times d}$ and perform cross-attention to extract category-related features from the spatial features using the Transformer decoder \cite{liu2021query2label}, where $C$ is the number of categories.
To alleviate the strong imbalance between positive and negative images in each category when taking multi-label classification, we follow the asymmetric loss \cite{ridnik2021asymmetric}, and refine it with a weighted asymmetric loss:
\begin{equation}
\begin{cases}
\mathcal{L}_+^{\text{mcls}} = \eta (1-p)^{\gamma_+} \text{log}(p) \\
\mathcal{L}_-^{\text{mcls}} = p^{\gamma_-} \text{log}(1-p)
\end{cases}
\end{equation}
where $p$ denotes the posterior probability with respect to a category, $\eta$ is the positive weight, $\gamma_+$ and $\gamma_-$ are the positive and negative focusing parameters. 

\textbf{Clustering decoder with prototypes for contrastive learning.} Given two image features $f_t$ and $f_s$ from two different augmentations of the same image, we compute their codes $q_t$ and $q_s$ by matching these features to a set of $K$ prototypes ${c_1, ..., c_K}$. We setup a “swapped” prediction problem with the following loss function:
\begin{equation}
\mathcal{L}^{\text{cl}}(f_t, f_s) = \textit{l}(f_t, q_s) + \textit{l}(f_s, q_t)
\end{equation}
Then, we define the $\textit{l}(f_t, q_s)$ as:
\begin{equation}
\textit{l}^{\text{cl}}(f_t, q_s) = - \sum_{k}{q_s^{(k)} \text{log}p_t^{(k)}}
\end{equation}
where $p_t^{(k)}=\text{softmax}({f_t}^{\text{T}}c_k/\tau)$, $\tau$ is a temperature parameter. The problem can be optimized by Sinkhorn-Knopp \cite{cuturi2013sinkhorn}. To avoid the collapse with the exponential operation, we adopt a truncated strategy by clamping the input tensor with a threshold of $T_{max}$.

\textbf{MIM decoder for reconstruction.} Inspired by the work in SimMIM \cite{xie2021simmim}, and use a learnable mask token vector to replace each masked patch. Image patches are the basic processing units of vision Transformers \cite{dosovitskiy2020image} \cite{liu2021swin}. it is convenient to apply the masking operation on patch-level that a patch is either fully visible or fully masked. For the model Swin\cite{liu2021swin}, we consider equivalent patch sizes of different resolution stages, $4 \times 4 \to 32 \times 32$, and adopt $32 \times 32$ by default, which is the patch size of the last stage. For ViT\cite{dosovitskiy2020image}, we adopt $32 \times 32$ as the default masked patch size. The reconstructive loss is defined as:
\begin{equation}
\mathcal{L}^{\text{mim}} = \frac{1}{\Omega (x)} \|y - x\|_1
\end{equation}
where $y \in \mathbf{R}^{H_0 \times W_0 \times 3}$ is the reconstruction of the input image $x$.

\textbf{Momentum distillation for label denoising.} As \cite{tencent-ml-images-2019} indicates, the annotated tags for most images in Open Images \cite{krasin2017openimages} are generated by machine, while only a few fraction of annotations are verified by humans. The noisy annotations are unavoidable and they are also included in the Tencent-ML dataset \cite{tencent-ml-images-2019}. To alleviate this, we propose to learn from pseudo-targets generated by the momentum model as that in \cite{li2021albef}. The momentum model is a continuously-evolving teacher which consists of exponential-moving-average (EMA) versions of the backbone and the Transformer decoder for multi-label classification. We train the base model such that its predictions match the ones from the momentum model. Specially, we take the vector of cosine similarities between image embedding and the corresponding label embeddings for momentum distillation. Here we define the cosine similarity as:
\begin{equation}
\mathcal{S}(g, Q_0) = Q_0 \otimes g
\end{equation}
where $g \in \mathbf{R}^d$ is the embedding vector learned from the above Transformer decoder, and $\otimes$ is the matrix product. Then, the distillation loss is defined as:
\begin{equation}
\begin{split}
\mathcal{L}^{\text{mom}} = E_{g, g'}(KL(\mathcal{S}(g, Q_0), \mathcal{S}(g', Q_0)) + \\
KL(\mathcal{S}(g', Q_0), \mathcal{S}(g, Q_0))) / 2
\end{split}
\end{equation}
where $g' \in \mathbf{R}^d$ is the embedding of momentum distillation, and $KL(\cdot)$ is the Kullback-Leibler (KL) divergence.

\subsection{\label{sec_3_3}Implementation details for pre-training}
Based on the above objectives, the full pre-training loss is as:
\begin{equation}
\mathcal{L} = \alpha_1 (\mathcal{L}_+^{\text{mcls}} + \mathcal{L}_-^{\text{mcls}}) + \alpha_2 \mathcal{L}^{\text{cl}} + \alpha_3 \mathcal{L}^{\text{mim}} + \alpha_4 \mathcal{L}^{\text{mom}}
\end{equation}
where $\alpha_1, \alpha_2, \alpha_3, \alpha_4$ are the weights for multi-label classification loss, contrastive loss, reconstruction loss and momentum distillation loss, and are set as 0.001, 0.02, 1.0 and 10.0 in our implementation, respectively. Besides, the parameters for multi-label classification, i.e. $\eta$, $\gamma_+$ and $\gamma_-$, are set as 10, 4, 1, respectively. The truncated threshold $T_{max}$ for Sinkhorn-Knopp is set as 10, and the momentum parameter for updating the momentum model is set as 0.995. For Swin-B \cite{liu2021swin} or ViT-B \cite{dosovitskiy2020image}, we pre-train the model for 30 epochs using a batch size of 1024 on 64 NVIDIA A100 GPUs. We use the AdamW \cite{loshchilov2017decoupled} optimizer with a weight decay of 0.05. The learning rate is warmed-up to 1e-4 in the first 5 epochs, and decayed to 1e-7 following a cosine schedule. During pre-training, we take random image crops of resolution 224 × 224 as input, and also apply Randaugment \cite{cubuk2020randaugment}.

\section{Experimental Results}
Generally, computer vision pipelines that employ self-supervised learning performs two tasks: a pretext task and a downstream task. The pre-training data with respect to the Tencent-ML dataset \cite{tencent-ml-images-2019} collects about 18 million images with 11,166 categories from existing well-known datasets, i.e., Open Images \cite{krasin2017openimages} and ImageNet \cite{deng2009imagenet}. To show the effectiveness of HVP-MTL as a foundation model, we conduct experiments on ImageNet-1K (IN-1K) classification \cite{deng2009imagenet}, COCO object detection \cite{lin2014microsoft}, and ADE20K \cite{zhou2019semantic} semantic segmentation, which are the most common downstream tasks in computer vision. We also provide comprehensive ablation studies on the effects of scaling backbones and each component of HVP-MTL.

\subsection{ImageNet-1K Classification}
ImageNet-1K was created by selecting a subset of 1.2M images from ImageNet dataset \cite{deng2009imagenet}, that belong to 1000 mutually exclusive classes. For fair comparison, we follow the training strategy in SimMIM, and train 100 epochs for all our models with the input size of $224 \times 224$. In Table \ref{tab:vitbin1k} and Table \ref{tab:swinbin1k}, we compare our proposed HVP-MTL with state-of-the-art (SOTA) pre-training methods, such as MoCo v3 \cite{chen2021empirical}, DINO \cite{caron2021emerging}, MAE \cite{he2021masked}, SimMIM \cite{xie2021simmim}, BEiT \cite{bao2021beit} and PeCo \cite{dong2021peco}, by measuring Top-1 accuracy on ImageNet-1K classification with the backbones of ViT-B and Swin-B, respectively. We also compare supervised pre-training models with the datasets of ImageNet-22K (IN-22K) and JFT-300M \cite{sun2017revisiting}. It shows that our method achieves the highest Top-1 accuracy with 84.2\% for ViT-B and 85.3\% for Swin-B, surpassing the supervised method with IN-1K by 2.4\% and 2.0\%, respectively. It is also worth noting that we achieve the same performance with the supervised method with JFT-300M for ViT-B. However, the later use a much larger dataset, and train more steps than ours. 

\begin{table}
\newcommand{\tabincell}[2]{\begin{tabular}{@{}#1@{}}#2\end{tabular}}
\centering
\begin{tabular}{ccccc} \\\hline
\tabincell{c}{Pre-train \\ Method} & \tabincell{c}{Pre-train \\ dataset} &  \tabincell{c}{Pre-train \\ epochs}  & \tabincell{c}{Input \\ size} & \tabincell{c}{Top-1 \\ accuracy} \\\hline
Supervised & - & - &  $224^2$ &81.8 \\
Supervised & IN-22K & 90 &  $384^2$ & 84.0  \\
Supervised & JFT-300M & 7 & $384^2$ & 84.2 \\
MoCo v3 & IN-1K &300 &  $224^2$ & 83.2 \\
DINO & IN-1K &300 &  $224^2$ & 82.8 \\
MAE & IN-1K &1600 &  $224^2$ & 83.6 \\
SimMIM & IN-1K &800 & $224^2$ & 83.8 \\
BEiT & IN-1K & 300 & $224^2$ & 82.8   \\
PeCo & IN-1K & 300 & $224^2$ &84.1 \\
HVT-MTL & Tencent-ML & 30 & $224^2$  &84.2 \\\hline
\end{tabular}
\caption{\label{tab:vitbin1k}Comparison of different pre-training methods on ImageNet-1K classification with the backbone of ViT-B.}
\end{table}

\begin{table}
\newcommand{\tabincell}[2]{\begin{tabular}{@{}#1@{}}#2\end{tabular}}
\centering
\begin{tabular}{ccccc} \\\hline
\tabincell{c}{Pre-train \\ Method} & \tabincell{c}{Pre-train \\ dataset} &  \tabincell{c}{Pre-train \\ epochs}  & \tabincell{c}{Input \\ size} & \tabincell{c}{Top-1 \\ accuracy} \\\hline
Supervised & - & - &  $224^2$ &83.3 \\
Supervised & IN-22K & 90 &  $384^2$ & 85.2  \\
SimMIM & IN-1K &800 & $224^2$ & 84.0 \\
HVT-MTL & Tencent-ML & 30 & $224^2$  &85.3 \\\hline
\end{tabular}
\caption{\label{tab:swinbin1k}Comparison of different pre-training methods on ImageNet-1K classification with the backbone of Swin-B.}
\end{table}

\subsection{COCO Object Detection}
Next, we evaluate different pre-training methods with Swin-B on COCO objection detection \cite{lin2014microsoft} with the Mask R-CNN framework \cite{he2017mask}. Specifically, we follow the fine-tuning strategy with 1$\times$ schedule, i.e. the 12 training epoch schedule, on the COCO training set. Table \ref{tab:swinbcoco} reports the results of different pre-training methods, such as DINO \cite{caron2021emerging}, PeCo \cite{dong2021peco} and supervised methods pre-trained on IN-1K and IN-22K. It shows that our proposed method outperforms all the counterparts. In details, our method outperforms the method on IN-22K by +1.0 box AP, and surpasses others by large margins. The promising results validate that large-scale supervised datasets are much valuable for visual representation, and can deliver useful information by transferring from classification tasks to object detection tasks.

\begin{table}
\newcommand{\tabincell}[2]{\begin{tabular}{@{}#1@{}}#2\end{tabular}}
\centering
\begin{tabular}{ccccc} \\\hline
\tabincell{c}{Pre-train \\ method} & \tabincell{c}{Pre-train \\ dataset} & $AP^b$  & $AP_{50}^b$ & $AP_{75}^b$ \\\hline
Supervised & IN-1K& 43.7 &  66.6 & 47.7 \\
Supervised & IN-22K& 46.9 &  68.8 & 51.6 \\
DINO & IN-1K & 43.2 &  66.2 & 47.6 \\
PeCo & IN-1K & 43.9 &  66.3 & 48.2 \\
HVT-MTL & Tencent-ML & 47.9 &  70.1 & 52.5 \\\hline
\end{tabular}
\caption{\label{tab:swinbcoco}Comparison of different pre-training methods on COCO object detection with the backbone of Swin-B.}
\end{table}

\subsection{ADE20K Semantic Segmentation}
We further investigate the capability of our method  for semantic segmentation on the ADE20K dataset \cite{zhou2019semantic} based on the backbone of Swin-B. Here, we employ Upernet \cite{xiao2018unified} as the basic framework. For fair comparison, we follow the previous work \cite{dong2021cswin}, and train Upernet with 160k iterations by setting batch size as 16. In Table \ref{tab:swinbade}, we report the results in terms of mIoU for different methods, such as DINO \cite{caron2021emerging}, BEiT \cite{bao2021beit}, PeCo \cite{dong2021peco} and supervised methods pre-trained on IN-1K and IN-22K. It can be seen that, our method also achieves the highest performance. Compared to the methods of purely self-supervised methods, the performance gain is very promising, and demonstrates the effectiveness of our pre-training method again.

\begin{table}
\newcommand{\tabincell}[2]{\begin{tabular}{@{}#1@{}}#2\end{tabular}}
\centering
\begin{tabular}{ccccc} \\\hline
\tabincell{c}{Pre-train \\ method} & \tabincell{c}{Pre-train \\ dataset} & mIoU \\\hline
Supervised & IN-1K& 48.0 \\
Supervised & IN-22K& 50.31 \\
DINO & IN-1K & 44.2 \\
BEiT & IN-1K & 45.7 \\
PeCo & IN-1K & 46.7 \\
HVT-MTL & Tencent-ML & 50.6 \\\hline
\end{tabular}
\caption{\label{tab:swinbade}Comparison of different pre-training methods on ADE20K semantic segmentation with the backbone of Swin-B.}
\end{table}

\subsection{Ablation Study}
To better understand HVP-MTL, we ablate each key component and evaluate the performance on ImageNet-1K classification based on ViT-B \cite{dosovitskiy2020image}. As explained above, there are four key designs in our methods, i.e., multi-label classification, MIM, contrastive learning for different image views, and momentum distillation for label denoising. As seen in Table \ref{tab:modelscale}, we observe relatively large performance drop on ImageNet classification by removing the multi-label classification or MIM task from our framework, indicating that learning with MIM and multi-label classification together is very crucial. 

\begin{table}
\newcommand{\tabincell}[2]{\begin{tabular}{@{}#1@{}}#2\end{tabular}}
\centering
\begin{tabular}{cc} \\\hline
Method & Top-1 accuracy \\\hline
HVT-MTL & 84.2 \\
w/o Multi-label classification & 83.6\\
w/o MIM & 83.8 \\
w/o Contrastive learning  & 84.1 \\
w/o Momentum distillation & 83.9  \\\hline
\end{tabular}
\caption{\label{tab:modelscale}Ablation study for pre-training using different strategies with ViT-B on ImageNet-1K classification.}
\end{table}

Then, we adopt Swin Transformer of different model sizes for pre-training experiments, including Swin-T, Swin-S, and Swin-B \cite{liu2021swin}. We train 30 epochs on the Tencent-ML dataset for all the pre-training tasks, and fine-tune with 100 epochs with the input size of $224 \times 224$. Table \ref{tab:modelscale} lists the results of our approach with different model sizes. With our pre-training, all of models achieve higher accuracy than their supervised counterparts. Specifically, models with larger size achieve more gains than smaller ones, showing good scalable charactistics for further improving the performance.

\begin{table}
\newcommand{\tabincell}[2]{\begin{tabular}{@{}#1@{}}#2\end{tabular}}
\centering
\begin{tabular}{ccccc} \\\hline
\tabincell{c}{Pre-train \\ method} & \tabincell{c}{Pre-train \\ dataset} & Swin-T & Swin-S & Swin-B \\\hline
supervised & - & 81.3 & 83.0 & 83.3\\
HVT-MTL & Tencent-ML & 81.6 & 83.7 & 84.2 \\\hline
\end{tabular}
\caption{\label{tab:modelscale}Ablation study for pre-training with backbones of different sizes on ImageNet-1K classification.}
\end{table}

\section{Conclusion}
This paper proposes HVP-MTL, a new framework for vision representation learning. HVP-MTL combines self-supervised and supervised visual tasks in a multi-task manner to cope with a few heuristic problems. We theoretically and experimentally verify the effectiveness of the proposed multi-task learning framework. Compared to existing methods, HVP-MTL offers better performance with the same vision models on multiple downstream tasks. For the future work, we plan to develop more powerful large models with good scaling performance for pre-training on large-scale multi-modal datasets, and employ more downstream tasks, such as depth/flow estimation, tracking, as well as additional vision and language tasks. In addition, the studies of adversarial attacks against pre-train models is also an interesting direction.

\bibliographystyle{alpha}
\bibliography{sample}

\end{document}